\documentclass[10pt,twocolumn,letterpaper]{article}

\usepackage{cvpr}
\usepackage{times}
\usepackage{epsfig}
\usepackage{graphicx}
\usepackage{amsmath}
\usepackage{commath}
\usepackage{amssymb}
\usepackage{enumerate,booktabs}
\usepackage{subfigure,color}
\usepackage[numbers,sort&compress]{natbib}
\usepackage{multirow}

\usepackage[pagebackref=true,breaklinks=true,letterpaper=true,colorlinks,bookmarks=false]{hyperref}

\cvprfinalcopy 

\def\cvprPaperID{2119} 

\begin{document}
\def\mA{\mathcal{A}}
\def\mB{\mathcal{B}}
\def\mC{\mathcal{C}}
\def\mD{\mathcal{D}}
\def\mE{\mathcal{E}}
\def\mF{\mathcal{F}}
\def\mG{\mathcal{G}}
\def\mH{\mathcal{H}}
\def\mI{\mathcal{I}}
\def\mJ{\mathcal{J}}
\def\mK{\mathcal{K}}
\def\mL{\mathcal{L}}
\def\mM{\mathcal{M}}
\def\mN{\mathcal{N}}
\def\mO{\mathcal{O}}
\def\mP{\mathcal{P}}
\def\mQ{\mathcal{Q}}
\def\mR{\mathcal{R}}
\def\mS{\mathcal{S}}
\def\mT{\mathcal{T}}
\def\mU{\mathcal{U}}
\def\mV{\mathcal{V}}
\def\mW{\mathcal{W}}
\def\mX{\mathcal{X}}
\def\mY{\mathcal{Y}}
\def\mZ{\mathcal{Z}}

\def\1n{\mathbf{1}_n}
\def\0{\mathbf{0}}
\def\1{\mathbf{1}}

\def\A{{\bf A}}
\def\B{{\bf B}}
\def\C{{\bf C}}
\def\D{{\bf D}}
\def\E{{\bf E}}
\def\F{{\bf F}}
\def\G{{\bf G}}
\def\H{{\bf H}}
\def\I{{\bf I}}
\def\J{{\bf J}}
\def\K{{\bf K}}
\def\L{{\bf L}}
\def\M{{\bf M}}
\def\N{{\bf N}}
\def\O{{\bf O}}
\def\P{{\bf P}}
\def\Q{{\bf Q}}
\def\R{{\bf R}}
\def\S{{\bf S}}
\def\T{{\bf T}}
\def\U{{\bf U}}
\def\V{{\bf V}}
\def\W{{\bf W}}
\def\X{{\bf X}}
\def\Y{{\bf Y}}
\def\Z{{\bf Z}}

\def\a{{\bf a}}
\def\b{{\bf b}}
\def\c{{\bf c}}
\def\d{{\bf d}}
\def\e{{\bf e}}
\def\f{{\bf f}}
\def\g{{\bf g}}
\def\h{{\bf h}}
\def\i{{\bf i}}
\def\j{{\bf j}}
\def\k{{\bf k}}
\def\l{{\bf l}}
\def\m{{\bf m}}
\def\n{{\bf n}}
\def\o{{\bf o}}
\def\p{{\bf p}}
\def\q{{\bf q}}
\def\r{{\bf r}}
\def\s{{\bf s}}
\def\t{{\bf t}}
\def\u{{\bf u}}
\def\v{{\bf v}}
\def\w{{\bf w}}
\def\x{{\bf x}}
\def\y{{\bf y}}
\def\z{{\bf z}}

\def\balpha{\mbox{\boldmath{$\alpha$}}}
\def\bbeta{\mbox{\boldmath{$\beta$}}}
\def\bdelta{\mbox{\boldmath{$\delta$}}}
\def\bgamma{\mbox{\boldmath{$\gamma$}}}
\def\blambda{\mbox{\boldmath{$\lambda$}}}
\def\bsigma{\mbox{\boldmath{$\sigma$}}}
\def\btheta{\mbox{\boldmath{$\theta$}}}
\def\bomega{\mbox{\boldmath{$\omega$}}}
\def\bxi{\mbox{\boldmath{$\xi$}}}
\def\bnu{\mbox{\boldmath{$\nu$}}}                                  
\def\bphi{\mbox{\boldmath{$\phi$}}}

\def\bDelta{\mbox{\boldmath{$\Delta$}}}
\def\bOmega{\mbox{\boldmath{$\Omega$}}}
\def\bPhi{\mbox{\boldmath{$\Phi$}}}
\def\bLambda{\mbox{\boldmath{$\Lambda$}}}
\def\bSigma{\mbox{\boldmath{$\Sigma$}}}
\def\bGamma{\mbox{\boldmath{$\Gamma$}}}

\newcommand{\myminimum}[1]{\mathop{\textrm{minimum}}_{#1}}
\newcommand{\mymaximum}[1]{\mathop{\textrm{maximum}}_{#1}}    
\newcommand{\mymin}[1]{\mathop{\textrm{minimize}}_{#1}}
\newcommand{\mymax}[1]{\mathop{\textrm{maximize}}_{#1}}
\newcommand{\mymins}[1]{\mathop{\textrm{min.}}_{#1}}
\newcommand{\mymaxs}[1]{\mathop{\textrm{max.}}_{#1}}  
\newcommand{\myargmin}[1]{\mathop{\textrm{argmin}}_{#1}} 
\newcommand{\myargmax}[1]{\mathop{\textrm{argmax}}_{#1}} 
\newcommand{\myst}{\textrm{s.t. }}

\newcommand{\denselist}{\itemsep -1pt}
\newcommand{\sparselist}{\itemsep 1pt}

\definecolor{pink}{rgb}{0.9,0.5,0.5}
\definecolor{purple}{rgb}{0.5, 0.4, 0.8}   
\definecolor{gray}{rgb}{0.3, 0.3, 0.3}
\definecolor{mygreen}{rgb}{0.2, 0.6, 0.2}

\newcommand{\cyan}[1]{\textcolor{cyan}{#1}}
\newcommand{\red}[1]{\textcolor{red}{#1}}  
\newcommand{\blue}[1]{\textcolor{blue}{#1}}
\newcommand{\magenta}[1]{\textcolor{magenta}{#1}}
\newcommand{\pink}[1]{\textcolor{pink}{#1}}
\newcommand{\green}[1]{\textcolor{green}{#1}} 
\newcommand{\gray}[1]{\textcolor{gray}{#1}}    
\newcommand{\mygreen}[1]{\textcolor{mygreen}{#1}}    
\newcommand{\purple}[1]{\textcolor{purple}{#1}}       

\definecolor{greena}{rgb}{0.4, 0.5, 0.1}
\newcommand{\greena}[1]{\textcolor{greena}{#1}}

\definecolor{bluea}{rgb}{0, 0.4, 0.6}
\newcommand{\bluea}[1]{\textcolor{bluea}{#1}}
\definecolor{reda}{rgb}{0.6, 0.2, 0.1}
\newcommand{\reda}[1]{\textcolor{reda}{#1}}

\def\changemargin#1#2{\list{}{\rightmargin#2\leftmargin#1}\item[]}
\let\endchangemargin=\endlist
                                               
\newcommand{\cm}[1]{}

\newcommand{\myheading}[1]{\vspace{0.05in}\noindent\textbf{#1}}

\title{Evolution-Preserving Dense Trajectory Descriptors}

\author{Yang Wang, Vinh Tran, Minh Hoai\\
Stony Brook University, Stony Brook, NY 11794, USA\\
{ \tt \small \{wang33, tquangvinh, minhhoai\}@cs.stonybrook.edu }
}
\maketitle

\begin{abstract}
Recently Trajectory-pooled Deep-learning Descriptors were shown to achieve state-of-the-art human action recognition results on a number of datasets. This paper improves their performance by applying rank pooling to each trajectory, encoding the temporal evolution of deep learning features computed along the trajectory. This leads to Evolution-Preserving Trajectory (EPT) descriptors, a novel type of video descriptor that significantly outperforms Trajectory-pooled Deep-learning Descriptors. EPT descriptors are defined based on dense trajectories, and they  provide complimentary benefits to video descriptors that are not based on trajectories. In particular, we show that the combination of EPT descriptors and VideoDarwin leads to state-of-the-art performance on Hollywood2 and UCF101 datasets.

\end{abstract}

\section{Introduction}

The ability to recognize human actions in video has many potential applications in a wide range of fields. 
Recognizing human actions in realistic videos, however, is difficult due to various challenges such as background clutter, self occlusion, and viewpoint variation. An effective and efficient approach to handle these challenges is to use local visual space-time descriptors, an approach that does not require non-trivial pre-processing steps such as recognition and segmentation. In particular, one of the best types of video descriptor is Dense Trajectory Descriptors (DTD)~\cite{Wang-Schmid-ICCV13,Wang-et-al-CVPR11}, which remain competitive even in the recent surge of deep-learning descriptors~\cite{Simonyan-Zisserman-NIPS14,tran2015learning, wang2015action}. In fact, most recent human action recognition methods~\cite{wang2015temporal, wang2015action, jain201515, Wang-Hoai-CVPR16, bilen2016dynamic, fernando2016discriminative, lev2015rnn} propose to combine Dense Trajectory Descriptors with deep learning features to obtain the best results. 

\begin{figure*}[t]
\begin{center}
\includegraphics[width=0.8\linewidth]{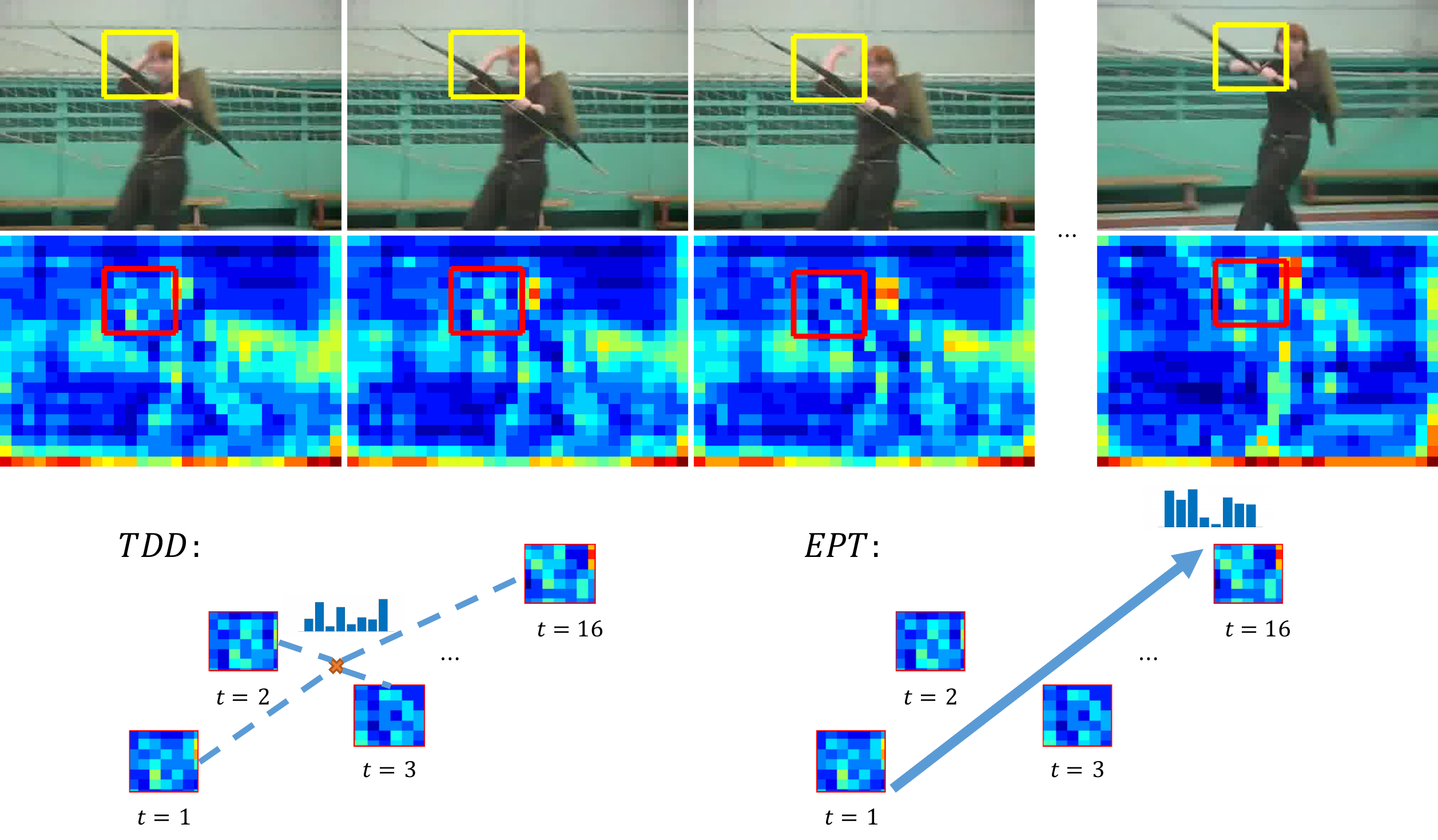}
\end{center}
\vskip -0in
   \caption{{\bf Illustration of the Evolution-Preserving Trajectory (EPT) Descriptors.} Given a trajectory, we consider the volume encapsulating that trajectory, and extract a sequence of deep-learning features along the spatiotemporal volume. We propose EPT, a novel descriptor that preserves the temporal evolution of appearance and motion along the trajectories. }
\label{fig:teaser}
\end{figure*}

Human actions, by definition, are caused by the motion of humans, and Dense Trajectory Descriptors exploit this fact. The power and success of Dense Trajectory Descriptors can be attributed to their ability to extract local descriptors along humans' motion. The key idea is to densely sample feature points  and track them in the video based on optical flow. Typically, the feature points are tracked for 15 frames, leading to trajectories with a temporal span of 15 frames. DTD explicitly estimates the camera motion and prunes away trajectories from the background (i.e., Improved DTD~\cite{Wang-Schmid-ICCV13}). It only retains trajectories from humans or objects of interest. 
For each trajectory, multiple descriptors are computed along the trajectory to capture shape, appearance,  and motion information. In particular, each trajectory leads to four feature vectors:  Histogram of Oriented Gradients~\cite{Dalal-Triggs-CVPR05}, Histogram of Optical Flow (HOF), Motion Boundary Histogram (MBH), and Trajectory Geometry (TG). Recently, \citet{wang2015action} propose to improve DTD by aggregating deep-learning descriptors (based on Two-stream CNN~\cite{Simonyan-Zisserman-NIPS14}) along each trajectory. This leads to the current state-of-the-art descriptors called  Trajectory-pooled Deep-learning Descriptors (TDD).

However, TDD ignores the temporal evolution of appearance and motion along the trajectories, which turns out to be important for describing human actions. Recent work of \citet{Fernando-etal-CVPR15} has demonstrated the benefits of modeling video-wide temporal evolution, and they proposed a method called Rank Pooling for capturing temporal evolution. Unlike mean and max pooling methods that ignore the sequential order of video frames in a video, rank pooling provides a way to aggregate information while respecting the temporal order of the video frames. \citet{Fernando-etal-CVPR15} used rank pooling to encode the evolution of  frame-based feature vectors, yielding impressive action recognition performance on a number of datasets. 

In this paper, we propose Evolution-Preserving Trajectory (EPT) descriptors, ones that combine the benefits of dense trajectories, deep-learning features, and rank pooling. As depicted in Figure \ref{fig:teaser}, EPT descriptors are based on dense trajectories, tagging along the apparent motion induced by human actions. EPT descriptors are based on deep-learning features, taking advantages of the  feature representation produced by Convolutional Neural Networks~\cite{LeCun-et-al-NC89,Krizhevsky-et-al-NIPS12,Simonyan-Zisserman-NIPS14} which have been trained on millions of images~\cite{Russakovsky-etal-IJCV15} and video frames~\cite{Soomro-et-al-TR12}. EPT descriptors also capture the evolution of appearance and motion along the trajectories.

EPT descriptors require performing rank pooling of deep-learning features along  densely populated trajectories. To avoid the prohibitive computational cost, we implement this procedure as follows. We first compute the video-wide convolution feature maps. We then use each trajectory's coordinates to access its corresponding deep-learning features. To preserve the temporal evolution along the trajectories, we use approximate rank pooling instead of the exact rank pooling. Approximate rank pooling is linear operation, which is much more efficient than the optimizing the Ranking Support Vector Machine~\cite{Fernando-etal-CVPR15}.

EPTs are local descriptors and they can be used in any standard recognition pipeline for action recognition. In this paper, we use EPT descriptors with Fisher Vector encoding~\cite{Perronnin-et-al-ECCV10} and Support Vector Machines~\cite{Vapnik-98}. We evaluate EPT descriptors on Hollywood2 and UCF101 datasets and find that they outperform the current state-of-the-art trajectories-based descriptors. When using EPT descriptors in conjunction with several pooling methods, the combined method outperforms the current state-of-the-art recognition performance on Hollywood 2 dataset.

\section{Related Work} \label{sec:related}

\myheading{Hand-crafted descriptors.} 
Feature descriptors are important for human action recognition, and much 
previous work has been devoted to the design of local features~\cite{Dollar-et-al-VSPETS05,Laptev-IJCV05,Wang-Schmid-ICCV13, Willems-et-al-ECCV08}, which are often robust to background clutter, self occlusion, and viewpoint variation in realistic video clips. There are various ways to extract informative regions from a video clip. For instance, Space Time Interest Points \cite{Laptev-IJCV05} use 3D Harris corner detector, while Cuboid \cite{Dollar-et-al-VSPETS05} uses temporal Gabor filters. In this paper, we choose the improved Dense Trajectories~\cite{Wang-Schmid-ICCV13} due to its good performance. Each trajectory leads to four feature vectors: Trajectory Geometry, Histograms of Oriented Gradient~\cite{Dalal-Triggs-CVPR05}, Histograms of Optical Flow, and Motion Boundary Histograms. These 4 descriptors are computed along the trajectory to capture information about shape, appearance, motion, and the change of motion, respectively.

\myheading{ Deep-learning descriptors. }
Deep learning has achieved great success in image based visual recognition tasks~\cite{Krizhevsky-et-al-NIPS12, simonyan2014very, Girshick-et-al-CVPR14, he2015deep}. There has been some successful attempts to develop deep architectures for action recognition~\cite{Simonyan-Zisserman-NIPS14,tran2015learning,wang2016temporal} as well. \citet{wang2015action} propose to improve dense trajectories descriptors by aggregating deep-learning descriptors (based on Two-stream CNN \cite{Simonyan-Zisserman-NIPS14}) along each trajectory. This leads to the current state-of-the-art descriptors called Trajectory-pooled Deep-learning Descriptors (TDD).

\myheading{ Rank Pooling. } 
In this paper, we propose to use rank pooling to preserve the temporal evolution along the trajectories. The concept of rank pooling is first introduced by \citet{Fernando-etal-CVPR15}. Their approach is referred to as VideoDarwin, which applies rank pooling to a sequence of video frames. Rank pooling has been applied to another extreme level of image pixels to produce dynamic images focusing on humans and objects of motion~\cite{bilen2016dynamic}.

\section{Evolution-Preserving Descriptors} \label{sec:ept}

In this section, we describe Evolution-Preserve Trajectory (EPT) descriptors, which are founded on dense trajectories, deep-learning features, and rank pooling. 

\subsection{Improved Dense Trajectories}

EPT descriptors are defined based on dense trajectories. To extract  dense trajectories, we  use the improved implementation of~\citet{Wang-Schmid-ICCV13} instead of the original dense trajectories~\cite{Wang-et-al-CVPR11}.
We customize the implementation of \citet{Wang-Schmid-ICCV13} with two modifications: (i) videos are normalized to have the height of 360 pixels, and (ii) frames are extracted at 25 fps. These modifications are added to standardize the feature extraction procedure across videos and datasets. They did not significantly alter the performance of the action recognition system.
Note that each trajectory is specified as a sequence of $L=16$ points in the video space, corresponding to a temporal span of 15 frames. The coordinates of these points on the trajectory are later used to pool deep convolutional features.

\subsection{Evolution Preserving }

\begin{figure}[t]
\begin{center}
\includegraphics[width=\linewidth]{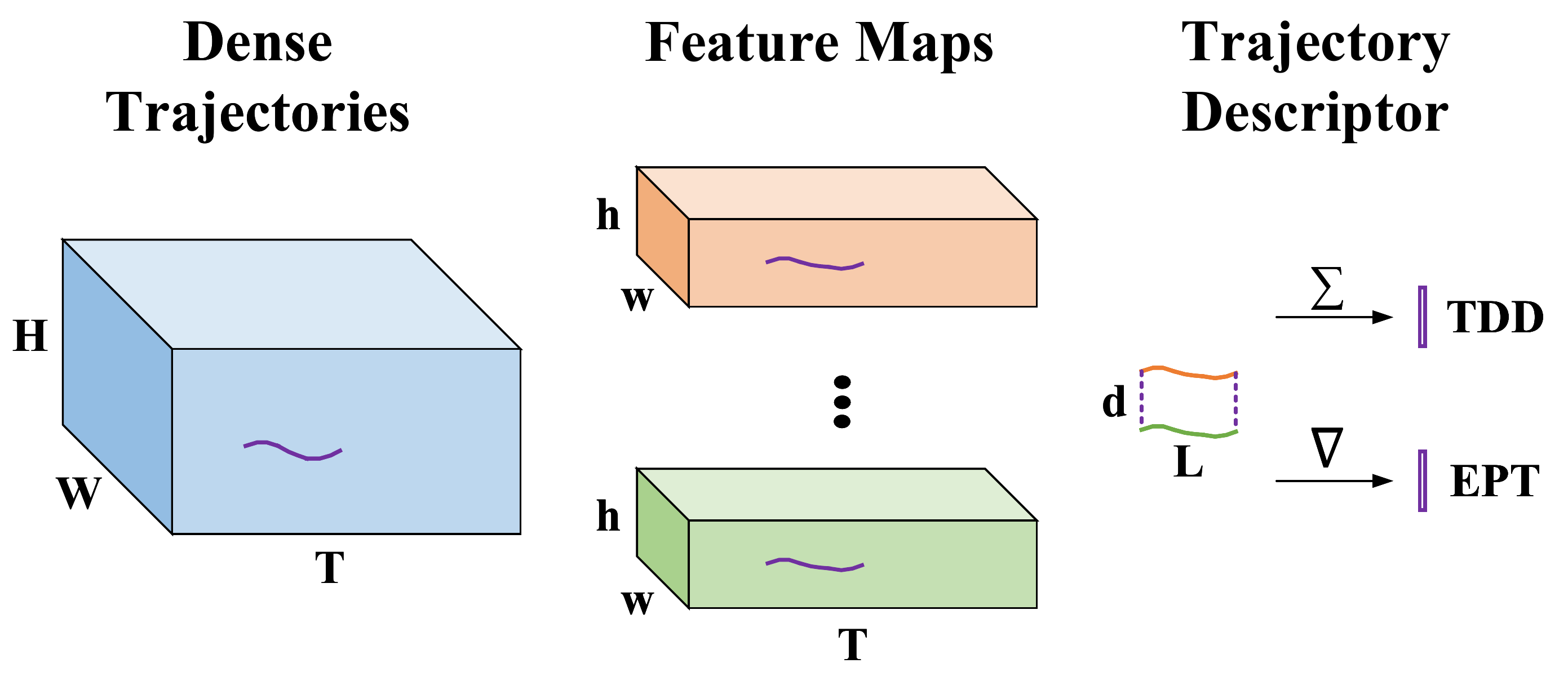}
\end{center}
\vskip -0in
   \caption{{\bf Illustration of computing deep-learning descriptors for trajectories.} Given a video, We extract the dense trajectories and the convolutional feature maps. Each trajectory is associated with a sequence of feature vectors $f_1, \cdots, f_L \in \mathbb{R}^d$, with $d$ being the number of CNN feature maps and $L$ the length of the trajectory. TDD applies Average Pooling to represent a trajectory. We propose to use Rank Pooling instead. }
\label{fig:TDD}
\end{figure}

The pipeline of computing deep-learning descriptors for trajectories is illustrated in Figure \ref{fig:TDD}. Given a video, we first extract thousands of trajectories as local interest points in the video space, and densely apply a CNN to extract the convolutional feature maps. Subsequently, for each trajectory, we consider a spatiotemporal volume encapsulating that trajectory; the length of the volume is the same as the length of the trajectory, and the width and and height of the volume are defined by the scale of the trajectory. Each trajectory-based volume is associated with a sequence of feature vectors $f_1, \cdots, f_L \in \mathbb{R}^d$, with $d$ being the number of CNN feature maps and $L$ the length of the trajectory. The current state-of-the-art trajectory-based method, TDD~\cite{wang2015action}, uses the average of $f_1, \cdots, f_L$ to represent the trajectory, i.e., to use $u_0=\frac{1}{L}\sum_{t=1}^{L}{f_t}$. 
Average pooling is simple approach, but it ignores the temporal progression of the feature vectors along the trajectory. 
In order to capture temporal evolution of the feature vectors, we propose to replace average pooling with rank pooling. Rank pooling obtains a vector vector $\mu^*$ to represent a sequence of feature vectors $f_1, \cdots, f_L$ by solving the following optimization problem:

\begin{align*}
& \mu^*  = \myargmin{\mu} \frac{\lambda}{2}\norm{\mu}^2 + \sum_{i<j}{ max\{0,1-\mu^T(F_j-F_i)\} }  \\
&\text{~where~} F_t =\left\lVert\sum_{\tau=1}^{t}{f_\tau}\right\rVert_2. 
\end{align*}
This is the original formulation of rank pooling, which is based on Ranking Support Vector Machine. The parameter vector $\mu$ is optimized to reflect the rank of the frames in the video. It  therefore aggregates relevant information over time and encodes the temporal progression. This is a convex optimization problem. The first term of the objective is a commonly used quadratic regularizer. The second term is a Hinge-loss, penalizing the pairs of frames $\{i,j\}$ that are incorrectly ranked. Note that a pair of frames $\{i,j\}$ is correctly ranked only if the ranking scores of the $L_2$-normalized accumulative feature vectors $\{F_i,F_j\}$ are separated by at least a unit margin, i.e. $\mu^T(F_j-F_i)>1$.

However, the above formulation requires solving an optimization problem for each trajectory, and this can be computationally prohibitive because there can be hundred thousand of trajectories for each video clip. Therefore, we propose to use approximate rank pooling instead~\cite{bilen2016dynamic}. The approximate rank pooling is derived by initializing $\mu$ as $\vec{0}$ and updating $\mu$ only once using gradient descent. As it turns out, the approximate rank pooling is essentially the average pooling of feature changes ($F_j-F_i$) over any time interval ($1\le i<j\le L$), which can be efficiently computed using a linear combination of the normalized accumulative feature vectors $\{F_t\}_{t=1}^{L}$: 

\begin{align}
\label{eq: approx-rank-pooling}
u_1=\sum_{1\le i<j\le L}{F_j-F_i}=\sum_{t=1}^{L}{(2t-L-1)F_t}.
\end{align}

\subsection{Convolutional Feature Maps } \label{subsec: conv_feature_map}

As shown in Figure \ref{fig:TDD}, CNN feature maps are required to extract deep-learning descriptors. The process of CNN feature map computation is  flexible with a variety of options, as listed in Table \ref{tab:featmap_options}. It is up to us to decide which CNN model and which convolutional layer to be used to extract the feature maps. The  size of the input video can also vary,  because the convolutional layers do not require a specific size for the input as in the case of fully-connected layers. There are several different feature map normalization methods as well. In the following, we provide a detailed discussion of these options.

\begin{table}[t]
\begin{center}
\begin{tabular}{ll}
\toprule
\textbf{Component} & \textbf{Options} \\
\midrule
\multirow{2}{*}{\textbf{model}} & spatial-stream VGG-M-2048 \\
                                & temporal-stream VGG-M-2048 \\
\midrule
\multirow{3}{*}{\textbf{scale}} & $360p \times 480p$ \\
                                & $240p \times 320p$ \\
                                & $180p \times 240p$ \\
\midrule
\multirow{2}{*}{\textbf{layer}} & spatial: conv4, conv5 \\
                       & temporal: conv3, conv4 \\
\midrule
\multirow{2}{*}{\textbf{normalization}} & in-voxel \\
                      & in-channel \\
\bottomrule
\end{tabular}
\end{center}
\caption{{\bf Possible options for the convolutional feature maps.} EPT descriptors are flexible, they can be paired with both spatial-and temporal-stream CNNs. One can choose the size of the input images, the layer of the CNN models to be used for extracting feature maps. The output feature maps can be normalized with different normalization methods.}
\label{tab:featmap_options}
\end{table}

\subsubsection{ Feature Map Extraction } 

We first discuss and specify some of the options used in the process of extracting convolutional feature maps.

\myheading{Two-stream CNN.} 
In principle, any kind of CNN architecture can be used to extract convolutional feature maps. In our implementation, we follow the original TDD paper and use a Two-stream CNN provided by~\citet{wang2015action}. The model is trained on UCF-101 dataset and contains both a spatial and a temporal CNN. The spatial CNN is based on VGG-M-2048 model and fine-tuned with single RGB frames (224$\times$224$\times$3); the temporal CNN has a similar structure, but its input is a volume of stacking optical flow fields (224$\times$224$\times$2F , $F=10$ is the number of stacking flows). 

For the purpose of extracting convolutional feature maps, we make two modifications to the original model. First, we make the model fully-convolutional by removing all the fully-connected layers. Second, we pad zero values  to each convolution layer's input with the size of the pad being half the size of the convolutional kernels. This is to avoid the boundary-shrinking effect of the convolutional kernels.

\myheading{Input Scale. } 
The size of the input RGB frames and optical flow images can vary, because we only use convolutional layers and convolutional layers do not require a fixed input size like the fully-connected layers. Having a larger input size means each neuron will look at a smaller neighborhood (perceptive field) and produce features at a finer scale. In our experiments, descriptors computed with larger input size usually has better performance. Combining descriptors at different input scales can improve the classification results. 

\myheading{Spatial Feature Maps. }
We extract frames from videos at 25fps and resize them into a predefined scale (e.g., $360p \times 480p$). Subsequently we feed individual frames into the pre-trained spatial CNN and take the output of the `conv4' or `conv5' layer as the spatial feature maps for each video, as suggested in the original TDD method~\cite{wang2015action}.

\myheading{Temporal Feature Maps. }
The process of computing the temporal feature maps is similar to the process of computing spatial feature maps, except we feed into the pre-trained temporal CNN with 10 consecutive optical flow images (with horizontal and vertical motion channels). The optical flow images are computed at 25fps using a GPU version of TVL1 algorithm, and subsequently resized into a predefined scale as well. For computing the temporal CNN descriptors, we used the `conv3' or `conv4' layer's output as the temporal feature maps.

\subsubsection{Feature Map Normalization} 

After the feature maps are extracted, we apply feature map normalization. Traditionally, normalization has been widely used for hand-crafted local descriptors such as HOG, HOF, and MBH. Here we discuss two methods that can be used to normalize convolutional feature maps: \textit{in-channel} normalization and \textit{in-voxel} normalization.

\begin{figure}[t]
\begin{center}
\includegraphics[width=\linewidth]{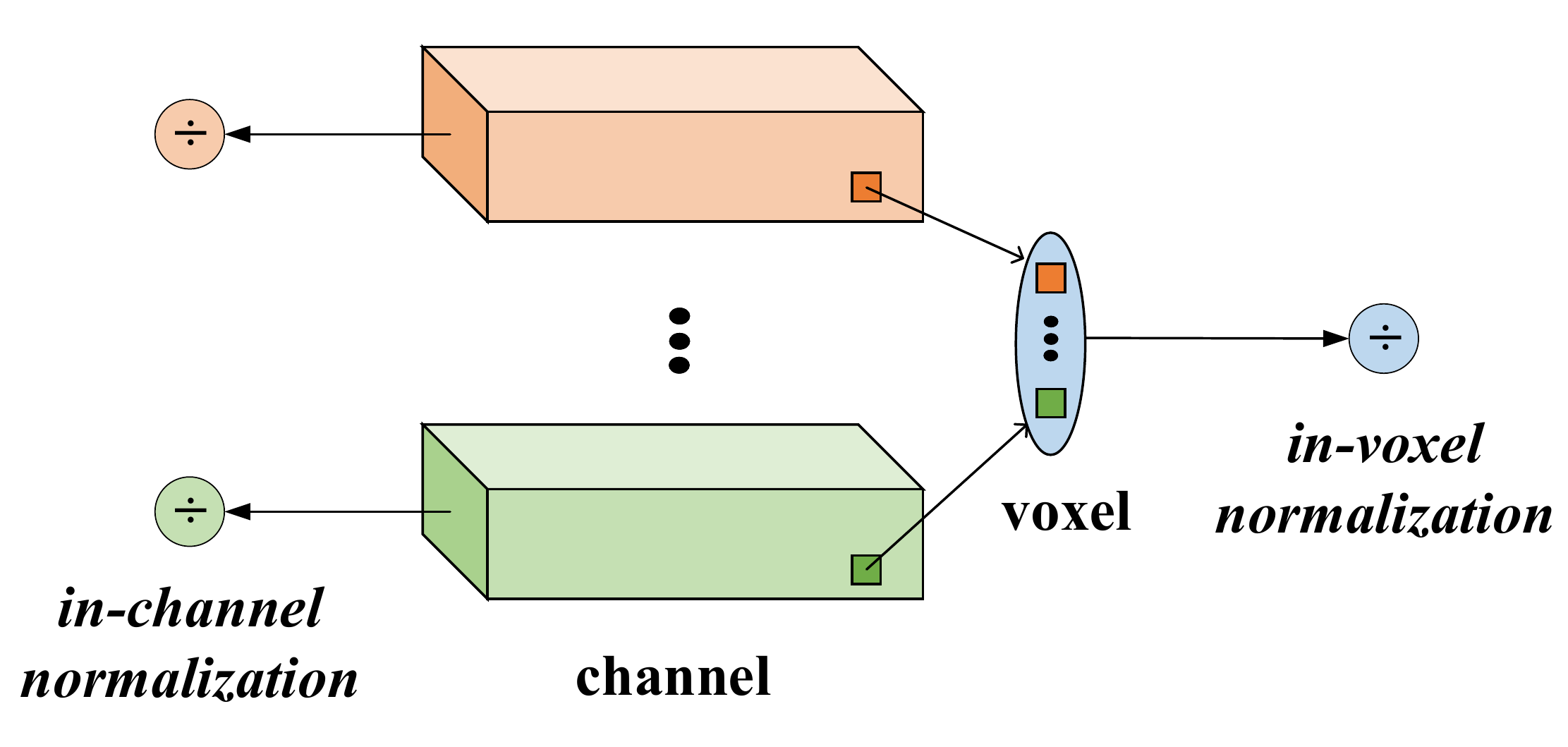}
\end{center}
\vskip -0in
   \caption{{\bf Illustration of Feature Map Normalization.} Left: \textit{in-channel} normalization is performed within each individual feature channel; Right: \textit{in-voxel} normalization is performed within each individual voxel.}
\label{fig:Normalize}
\end{figure}

\myheading{In-channel normalization. } 
Figure \ref{fig:Normalize} (left) illustrates the in-channel normalization method, which is defined as $L_\infty$ normalization of each individual feature channel across the video's spatiotemporal volume. This normalization is to ensure each feature channel ranges in the same interval, and thus contributes equally to final representation.

\myheading{In-voxel normalization. }
Figure \ref{fig:Normalize} (right) also illustrates the in-voxel normalization method. It is defined as $L_\infty$ normalization of each individual voxel across the feature channels. This ensures each voxel ranges in the same interval, and consequently makes equal contribution to the final representation.

\subsection{ Fisher Vector encoding }

Finally, after computing the local descriptors, we use Fisher Vector~\cite{Perronnin-et-al-ECCV10} to represent an entire video. A Fisher Vector encodes both first and second order statistics between the feature descriptors and a Gaussian Mixture Model (GMM). In~\cite{Wang-Schmid-ICCV13}, Fisher Vector shows an improved performance over bag of features for action classification. 

For each type of descriptors, we first sample a subset of 1,000,000 data points, and use PCA to de-correlate the descriptors and reduce the dimension to $D$ ($D=64$ for CNN-based descriptors; $D$ is half of the original dimension for iDT descriptors), and then train a GMM with $K=256$ mixtures. Finally, each video is represented with a $2KD$-dim Fisher Vector, which is subsequently power ($\alpha=0.5$) normalized and $L_2$ normalized, as in~\cite{Perronnin-et-al-ECCV10,Wang-Schmid-ICCV13}.

After encoding the Fisher Vectors, we train one-vs-all SVMs~\cite{Vapnik-98} for action classification. After learning the classifiers, given a test video, we compute the probability of each action by normalizing the scores across actions with a softmax function~\cite{Wang-Hoai-CVPR16}. In order to fuse different descriptor types, we average their corresponding linear kernels, which is equivalent to concatenating their corresponding Fisher Vectors.

\begin{figure}[t]
\begin{center}
\includegraphics[width=0.99\linewidth]{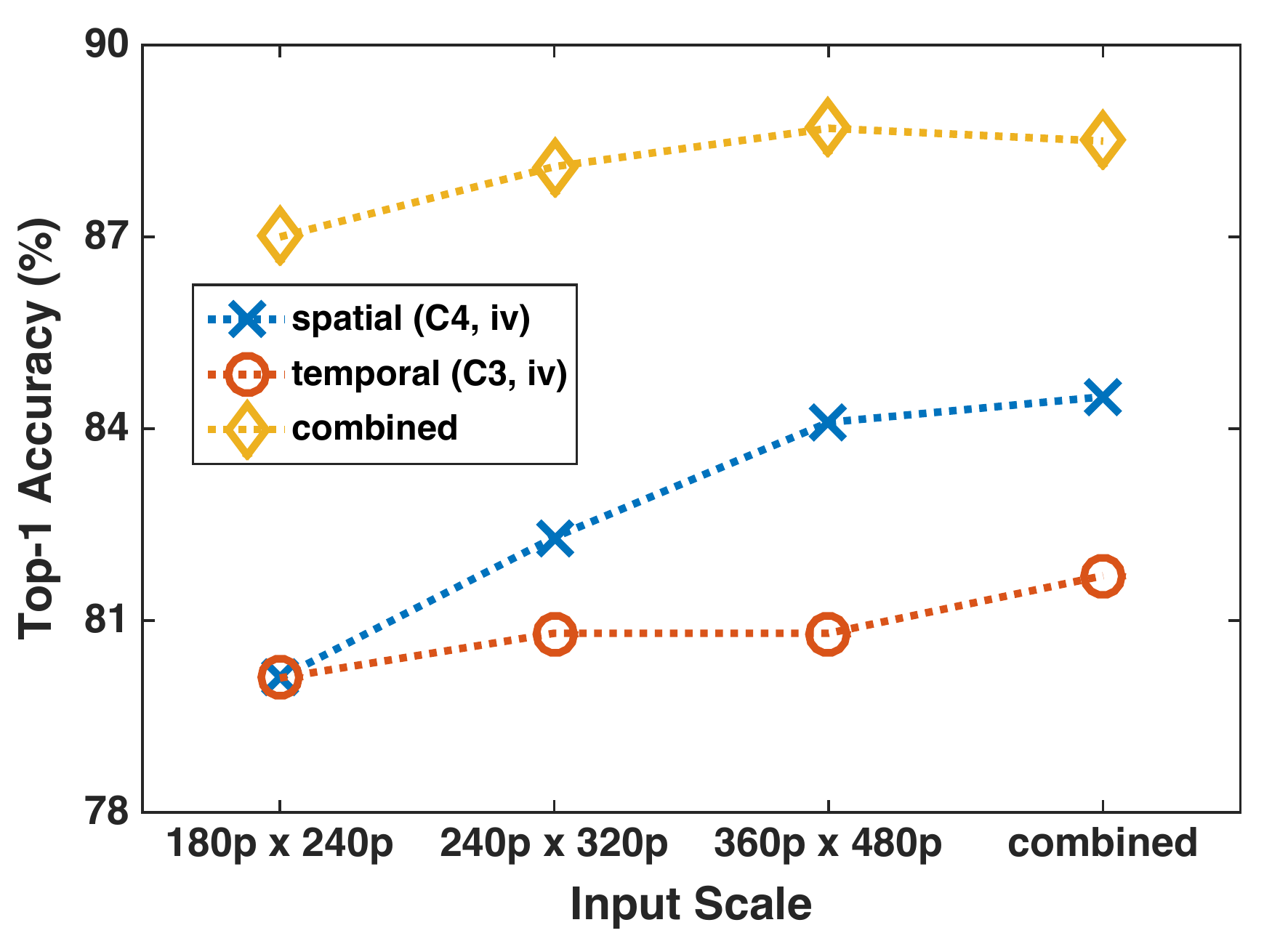}
\end{center}
\vskip -0.1in
   \caption{{\bf The performance of EPT descriptors at different input scales.} On UCF101 dataset (split-1), the EPT descriptors computed at a larger input size usually outperforms those computed at a smaller input size. Combining descriptors at multiple scales achieve the best result, for both the spatial and temporal streams.}
\label{fig:scale}
\end{figure}

\begin{figure}[t]
\begin{center}
\includegraphics[width=0.99\linewidth]{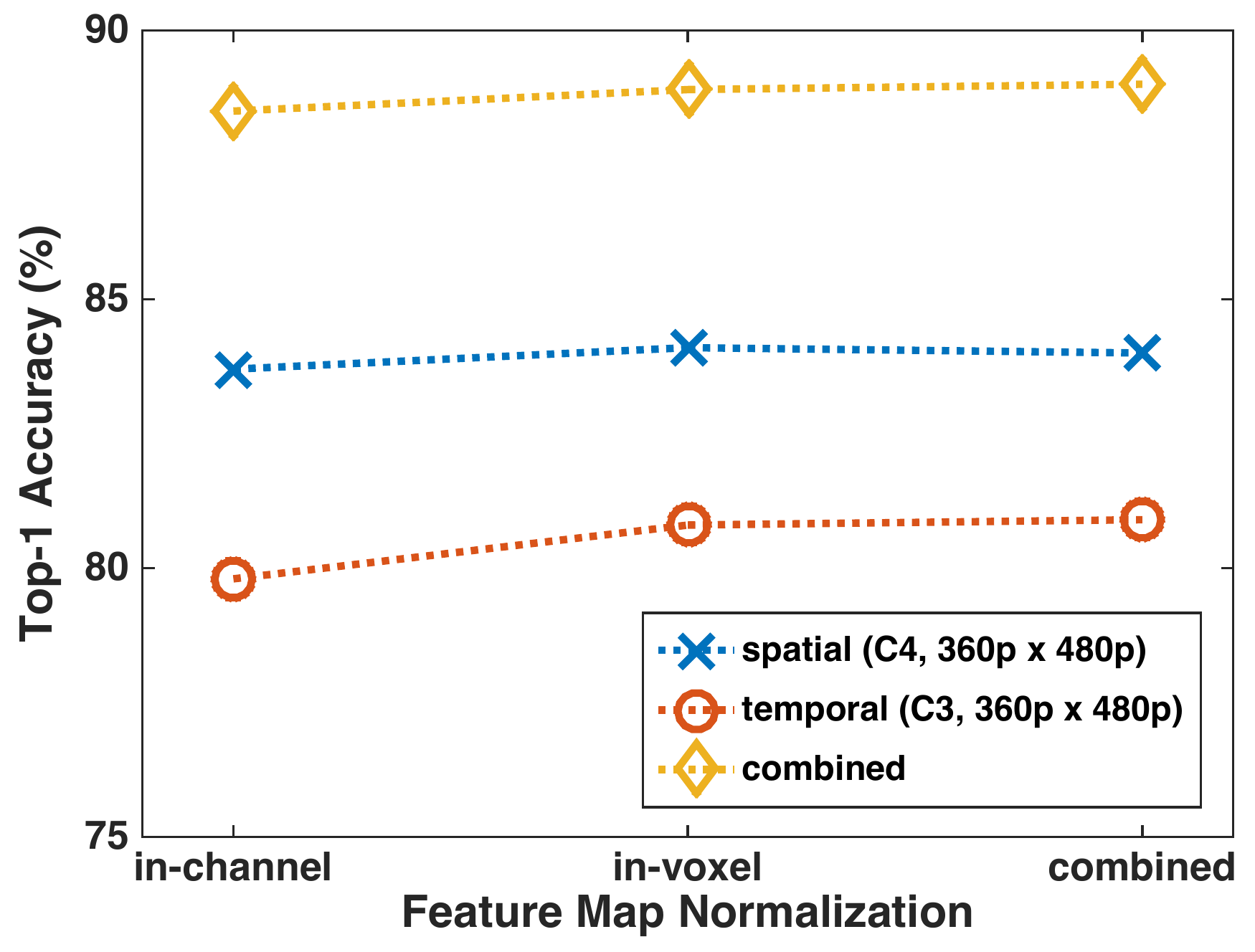}
\end{center}
\vskip -0.1in
   \caption{{\bf Comparing different normalization methods.} This shows the performance of EPT descriptors on the UCF101 dataset (split-1). In-voxel normalization consistently outperforms in-channel normalization. Using both normalization methods does not lead to better performance than using in-voxel normalization alone.  We suggest to use in-voxel normalization only.}
\label{fig:norm}
\end{figure}

\section{ Experiments } \label{sec:experiments}

\subsection{ Datasets }
We evaluate the performance of EPT descriptors using two datasets: Hollywood2~\cite{Marszalek-et-al-CVPR09} and UCF101~\cite{Soomro-et-al-TR12}. The Hollywood2 dataset has 12 action classes and contains 1707 video clips collected from 69 different Hollywood movies. The videos are split into a training set of 823 videos and a testing set of 884 videos. The training and testing videos come from different movies. We augment the training set with horizontally flipped training videos, and report the mean Average Precision as the performance measure on Hollywood2.
The UCF101 dataset contains 101 action categories with 13,320 realistic action videos collected from YouTube. Each category has at least 100 video clips. The dataset has three different training/test splits. We follow the evaluation scheme of the THUMOS13 challenge~\cite{jiang2014thumos} and adopt the top-1 accuracy as the evaluation metric for each training/test split in UCF101.

\subsection{Exploratory Studies}

Table \ref{tab:featmap_options} lists 4 components that can be configured in our pipeline: \textit{model}, \textit{layer}, \textit{scale}, \textit{normalization}. However, the \textit{model} and \textit{layer} are not subject to change, because we use the same models and layers in the original TDD paper~\cite{wang2015action}, to make the TDD baselines as strong as possible. As for the \textit{scale} and \textit{normalization}, we analyze their effect below.\\

\myheading{Input Scale.} We first investigate the performance of EPT descriptors computed at different input scales on the UCF101 dataset (split-1). In this experiment, we fix the `conv4' activations (after RELU) as the spatial feature maps and `conv3' activations (after RELU) as the temporal feature maps. All feature  maps are normalized using in-voxel normalization. We consider three input sizes, $180p \times 240p$, $240p\times 320p$, and $360p \times 480p$, and the results are shown in Figure \ref{fig:scale}. As can be observed, the EPT descriptors computed at a larger input size  performs better than those computed at a smaller input size. This might be attributed to having smaller receptive field. With a larger input size, each neuron looks at a smaller neighborhood (perceptive field) and consequently outputs the descriptors at a finer scale, leading to better performance. Of course, this is only true up to a certain extent. If the input size becomes extremely large, the neurons only observe tiny regions that are semantically meaningless. Also, a larger input size requires more computational power and memory capacity.  In Figure \ref{fig:scale}, we also notice that combining descriptors at multiple scales can further improve the performance on both the spatial and temporal streams.

\myheading{Normalization method.} The normalization methods for convolutional feature maps is another important factor to investigate. In this experiment, the spatial feature maps (`conv4') and the temporal feature maps (`conv3') are both computed at input scale of $360p \times 480p$. The feature maps are either normalized using in-channel normalization or in-voxel normalization. Figure \ref{fig:norm} compares the two normalization methods on the UCF101 dataset (split-1). In general, in-voxel normalization is slightly better than  in-channel normalization for EPT descriptors. The reason might be that, with in-voxel normalization, each voxel is equally represented in the same interval and therefore making the rank pooling more effective in terms of discerning the changes among adjacent voxels. Figure \ref{fig:norm} also suggests that the combination of the two normalization methods does not lead to significant improvement over the use of in-voxel normalization alone. Therefore, we only use in-voxel normalization in the following experiments.

\myheading{Chosen configuration.} Equipped with the knowledge learned from these exploratory experiments, we specify some options in Table \ref{tab:exp_setup} for the evaluation of EPT descriptors. In order to extract convolutional feature maps, we apply the Two-stream CNN provided by \citet{wang2015action}. The `conv4' activations (after RELU) of the spatial CNN are taken as the spatial feature maps. For temporal CNN, we consider the `conv3' activations as the temporal feature maps. The input size of both RGB frames and optical flow images is set to be $360p \times 480p$. All convolutional feature maps are normalized using in-voxel normalization.

\begin{table}[t]
\begin{center}
\begin{tabular}{ll}
\toprule
\textbf{Component} \hspace{5ex} & \textbf{Used Configuration} \\
\midrule
\textbf{model} & Two-stream VGG-M-2048 \\
\midrule
\textbf{scale} & $360p \times 480p$ \\
\midrule
\multirow{2}{*}{\textbf{layer}} & spatial: conv4 \\
                                & temporal: conv3 \\
\midrule
\textbf{normalization} & in-voxel \\
\bottomrule
\end{tabular}
\end{center}
\caption{{Chosen configuration for the CNN feature maps used by EPT descriptors.} This configuration was determined based on our exploratory studies.}
\label{tab:exp_setup}
\end{table}

\subsection{Evaluation of EPT Descriptors}

\begin{figure}[t]
\begin{center}
\includegraphics[width=0.9\linewidth]{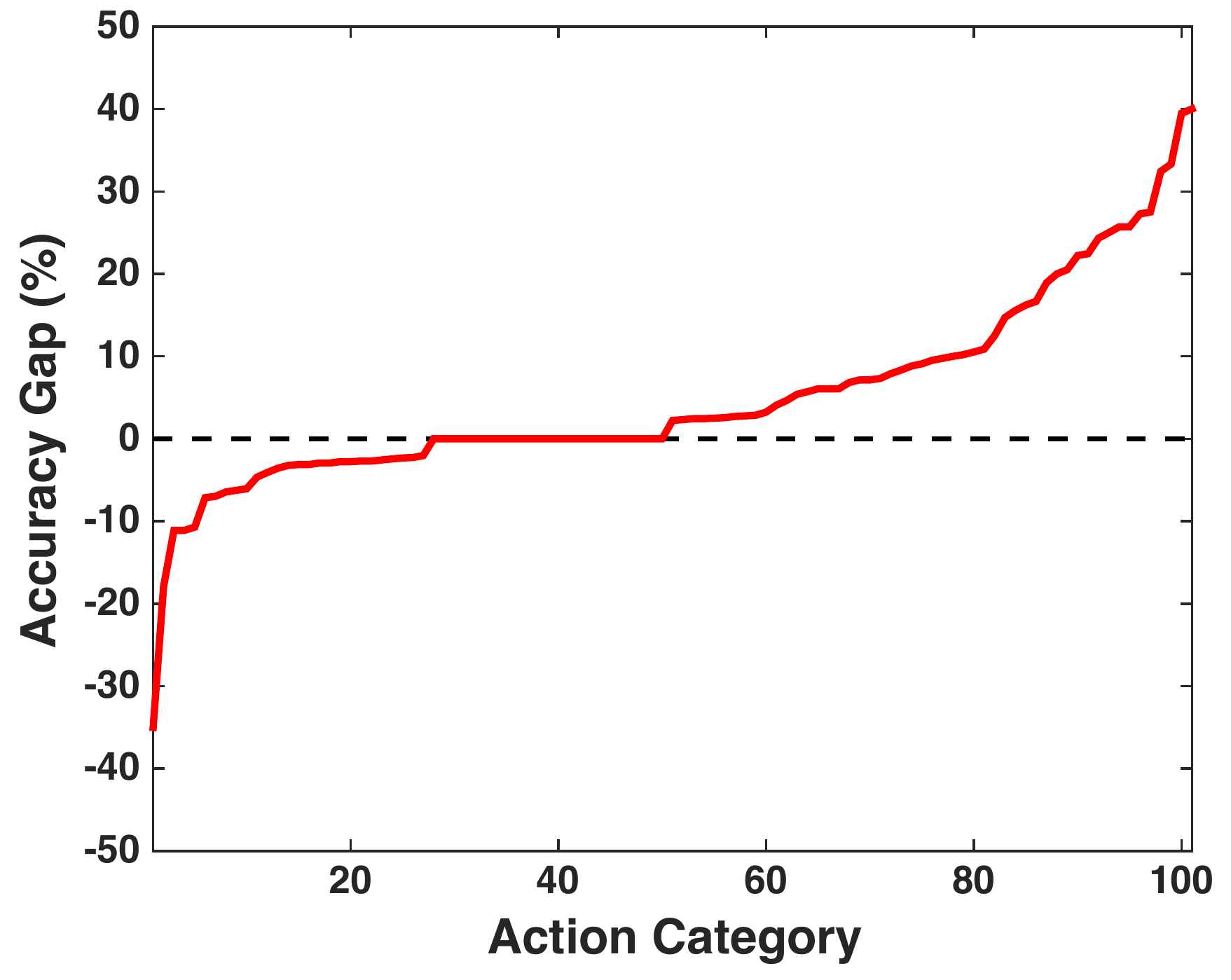}
\end{center}
\vskip -0.1in
   \caption{{\bf Accuracy gap between spatial-EPT and spatial-TDD for 101 classes of the UCF101 dataset (split-1)} A positive gap indicates the improvement in classification accuracy from spatial-TDD to spatial-EPT. Most of the classes have positive accuracy gaps. For some actions, the improvement can be as high as 40\%.}
\label{fig:acc_gap}
\end{figure}

We evaluate the performance of the proposed EPT descriptors on both Hollywood2 and UCF101 dataset (split-1), and the experiment results are summarized in Table \ref{tab: traj_pool_result}. It's worth noting that, as did in~\cite{Fernando-etal-CVPR15,fernando2016discriminative}, we apply rank-pooling in both forward and backward directions. The forward and backward EPT descriptors are then fused by concatenating their corresponding Fisher vectors. As can be observed, compared to the TDD descriptors, EPT descriptors significantly improve the recognition performance on both the spatial and temporal streams, which indicates the importance of preserving the temporal evolution of appearance and motion along the trajectories. The advantage is most evident on the spatial stream, where EPT outperforms TDD by 7\% on both dataset (from 45.5\% to 52.8\% on Hollywood2; from 77.5\% to 84.1\% on UCF101). 
Figure \ref{fig:acc_gap} depicts the accuracy gap, the difference between the accuracy of spatial-TDD and the accuracy of spatial-EPT for 101 classes on the UCF101 dataset. As can be seen, most of the classes have a positive accuracy gap, indicating the effectiveness of rank pooling over average pooling. For some actions, the improvement can be as high as 40\%. For the temporal stream, the EPT descriptors are also better than the TDD descriptors, even though the temporal CNN already takes as input a stack of optical flow images and models the temporal evolution of human motions. 

\begin{figure*}[t]
\begin{center}
\includegraphics[width=0.9\linewidth]{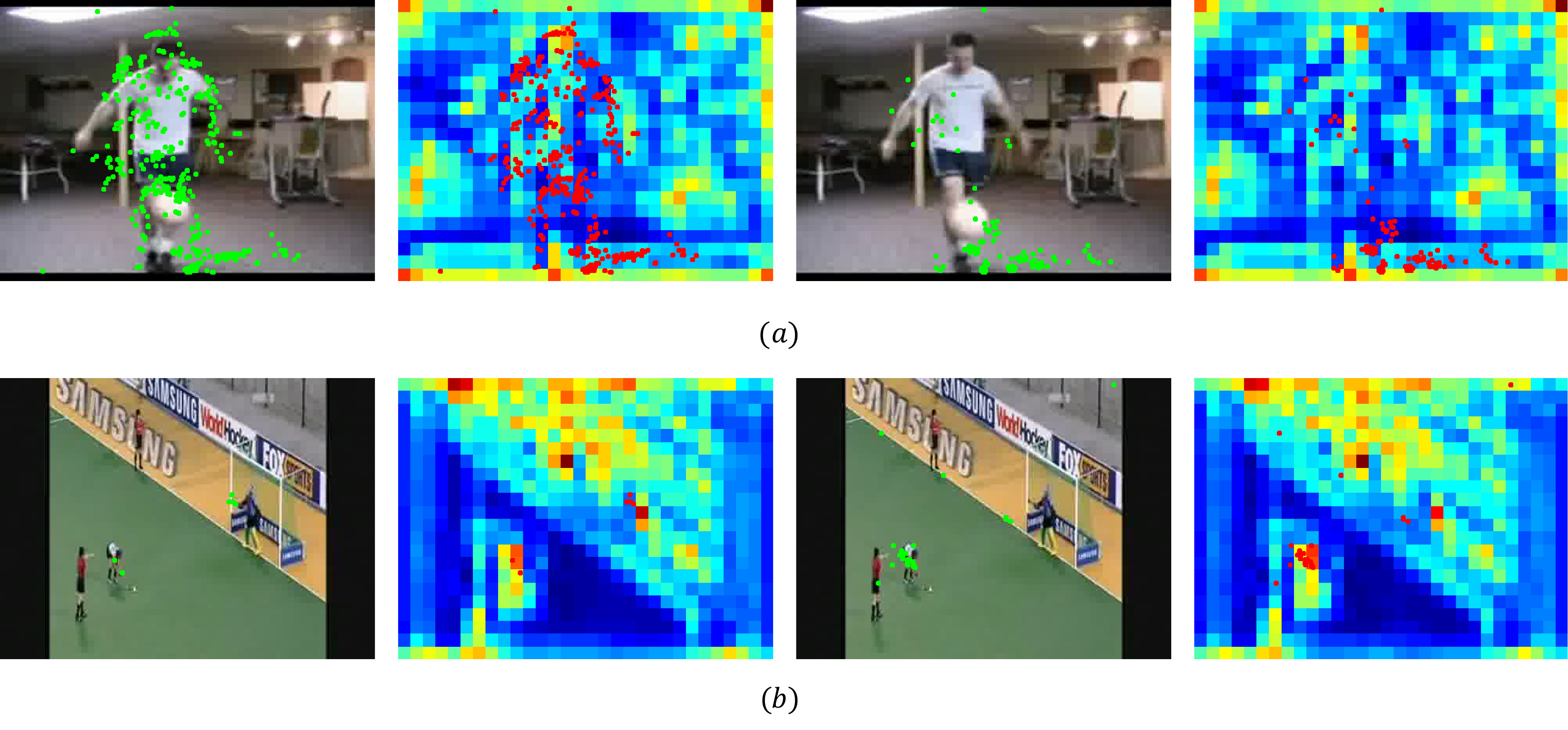}
\end{center}
\vskip -0.1in
   \caption{{\bf Success case (a) and failure case (b) for the spatial EPT.}  Clip (b) is correctly classified by the spatial TDD, but misclassified by the spatial EPT. The amount of motion and trajectories in Clip (b) is relatively small. Because the appearance change along the trajectory is small, the spatial EPT descriptor becomes close to zero, failing to describe the static scenes around the trajectory. This is a drawback of using rank pooling over average pooling. This problem will be addressed in our future work. }
\label{fig:success_fail}
\end{figure*}

In Figure \ref{fig:success_fail}, we show two video clips. Clip (a) is misclassified by the spatial TDD, but correctly classified by the proposed spatial EPT. Clip (b), on the other hand, is correctly classified by the spatial TDD, but misclassified by our spatial EPT. As can be seen, the amount of motion and tracked trajectories in Clip (b) is relatively small. Because the appearance change along the trajectory is small, the spatial EPT descriptor becomes close to zero, failing to describe the static scenes around the trajectory. This is a drawback of rank pooling over average pooling. Currently we address this issue by fusing EPT with TDD. In future work, we will seek a descriptor that encodes both the static appearance and the appearance change along the trajectories.

\begin{table}[t]
\begin{center}
\begin{tabular}{l|l|ccc}
\toprule
\multirow{2}{*}{Dataset} & Feature & \multicolumn{3}{c}{Trajectory-level Pooling} \\
 & Maps & {\small \textbf{TDD}} & {\small \textbf{EPT}} & {\small \textbf{Improve}} \\
\midrule
\multirow{3}{*}{\small Hollywood2} & spatial   & 45.5 & \textbf{52.8} & 7.3 \\
                            & temporal  & 61.4 & \textbf{63.3} & 1.9 \\
                            & 2-stream  & 64.8 & \textbf{67.1} & 2.3 \\
\midrule
\multirow{3}{*}{\small UCF101} & spatial & 77.5 & \textbf{84.1} & 6.6 \\
                        & temporal & 77.9 & \textbf{80.8} & 2.9 \\
                        & 2-stream & 86.1 & \textbf{88.7} & 2.6  \\
\bottomrule
\end{tabular}
\end{center}
\caption{{\bf Comparing EPT and TDD descriptors on Hollywood2 and UCF101 (split-1).}  EPT descriptors significantly improve the recognition performance on both the spatial and temporal streams, which indicates the importance of preserving the temporal evolution of appearance and motion along the trajectories. The advantage is most evident on the spatial stream.}
\label{tab: traj_pool_result}
\end{table}

\subsection{Using EPT descriptors with different video-level pooling methods}

After verifying the advantage of EPT descriptors over TDD descriptors on the trajectory level, we further investigate the performance of EPT descriptors combined with different video-level pooling methods. 

Each trajectory has a temporal span of 15 frames, and we assign each trajectory to its middle frame (the $8^{th}$ frame). Each frame is therefore associated with a set of trajectories, and we can compute an unnormalized Fisher Vector for each frame. Consequently, a video can be represented as a sequence of frame-wise unnormalized Fisher Vectors $\phi_1, \cdot\cdot\cdot, \phi_T$. In order to compress the information from the sequence into a single feature vector, we can apply many different video-level pooling methods. Up to now, we have been using average pooling to represent an entire video as $\phi=norm(\sum_{t=1}^{T}{\phi_t})$. As aforementioned, we set the $norm$ function to be power normalization ($\alpha=0.5$) followed by $L_2$ normalization. We can also apply video-level rank pooling, i.e., VideoDarwin~\cite{Fernando-etal-CVPR15}). Aside from average pooling (AP) and rank pooling (RP), we also propose a simple and effective pooling method called hierarchical average pooling (HAP). To obtain the final feature vector using hierarchical average pooling, we simply run a sliding window (size: 20, stride: 1) and perform average pooling and normalization within each window, and subsequently perform average pooling over the entire video again. We suggest to use HAP instead of AP, because HAP consistently outperforms AP in our experiments.

Table \ref{tab: video_pool_result}  compares the performance of EPT descriptors combined with different video-level pooling methods on Hollywood2 and UCF101  datasets. As can be observed, both the hierarchical average pooling (HAP) and the rank pooling (RP) outperform the ordinary average pooling (AP), and their fusion can further improve the classification results. Interestingly, the improvement of rank pooling over average pooling at the video level is more evident for the temporal stream, whereas at the trajectory level the improvement is higher for the spatial stream. On the Hollywood2 dataset, we notice that the fusion between EPT and DTD descriptors achieves significant improvement. On the UCF101 dataset, however, the combination with DTD does not lead to much better performance.

\begin{table}[t]
\begin{center}
\begin{tabular}{l|l|cccc}
\toprule
\multirow{2}{*}{Dataset} & \multirow{2}{*}{\textbf{EPT}} & \multicolumn{4}{c}{Video-level Pooling} \\
 & & {\small \textbf{AP}} & {\small \textbf{HAP}} & {\small \textbf{RP}} & {\small \textbf{HAP+RP}} \\
\midrule
\multirow{4}{*}{\small Hollywood2} & spatial   & 52.8 & 54.1 & 55.9 & \textbf{56.2} \\
                                   & temporal  & 63.3 & 67.0 & 68.3 & \textbf{68.8}  \\
                                   & 2-stream  & 67.1 & 69.6 & \textbf{71.4} & 71.2 \\
                                   \cline{2-6}  \rule{0pt}{2.6ex}
                                   & + DTD    & 72.5 & 74.6 & \textbf{76.3} & \textbf{76.3} \\
\midrule
\multirow{4}{*}{\small UCF101} & spatial  & 84.1 & \textbf{84.5} & 83.6 & 84.1 \\
                               & temporal & 80.8 & 82.6 & 82.1 & \textbf{82.7} \\
                               & 2-stream & 88.7 & \textbf{89.5} & 88.8 & 89.2 \\
                               \cline{2-6}  \rule{0pt}{2.6ex}
                               & + DTD    & 88.5 & \textbf{89.5} & 89.2 & \textbf{89.5} \\
\bottomrule
\end{tabular}
\end{center}
\caption{{\bf Comparing different video-level pooling methods for EPT descriptors.} Hierarchical average pooling (HAP) and rank pooling (RP) outperform average pooling (AP) video-wide, and their fusion can further improve the results. The improvement of video-wide rank pooling is more evident for the temporal stream. On Hollywood2, the fusion of EPT and DTD descriptors achieves significant improvement.}
\label{tab: video_pool_result}
\end{table}

\subsection{Comparison to the state-of-the-art}
In order to achieve the best recognition performance, we compute EPT descriptors with multiple convolutional layers (spatial: `conv4'+`conv5'; temporal: `conv3'+`conv4') and input scales ($180p \times 240p$, $240p \times 320p$, $360p \times 480p$). We also fuse EPT with DTD descriptors and apply the VideoDarwin technique. Table \ref{tab:state-of-the-art} compares our recognition performance with several other state-of-the-art methods on Hollywood2 and UCF101. On the Hollywood2 dataset, our method outperforms previous state-of-the-art by 2\%. On the UCF101 dataset, we achieve performance on par with the Temporal Segment CNN~\cite{wang2016temporal}, which is a very-deep version of the Two-stream CNN based on the VGG-VeryDeep-16 model~\cite{simonyan2014very} with three input modalities: RGB, optical flow, and warped optical flow. In future work, we will use these new models to extract convolutional feature maps and seek further improvement.

\begin{table}[t]
\begin{center}
\begin{tabular}{lcc}
\toprule
 & Hollywood2 & UCF101 \\
\midrule
DTD + FV~\cite{Wang-Schmid-ICCV13}     & 64.7 & 85.9 \\
SSD + RCS~\cite{Hoai-Zisserman-ACCV14}    & 73.6 & --   \\
VideoDarwin~\cite{Fernando-etal-CVPR15}  & 73.7 & --   \\
Two Stream~\cite{Simonyan-Zisserman-NIPS14}   & -- & 88.0 \\
DIN + DTD~\cite{bilen2016dynamic}    & --   & 89.1 \\
C3D + DTD~\cite{tran2015learning}    & --   & 90.4 \\
HRP + DTD~\cite{fernando2016discriminative}    & 76.7 & 91.4 \\
TDD + DTD~\cite{wang2015action}    & --   & 91.5 \\
RNN-FV +DTD~\cite{lev2015rnn}    & --  & 94.1  \\
TSN~\cite{wang2016temporal}      & -- & \textbf{94.2} \\
\midrule
Our method  & \textbf{78.6} & 92.3 \\
\bottomrule
\end{tabular}
\end{center}
\caption{{\bf Comparison with other state-of-the-art methods.} We compute EPT descriptors with multiple convolutional layers and input scales to preserve short-term evolution, and fuse them with DTD descriptors. We also apply VideoDarwin to capture long-term dynamics. On Hollywood2, our method outperforms previous state-of-the-art by 2\%. On UCF101, we achieve performance on par with the Temporal Segment CNN, which is a very-deep version of the Two-stream CNN. In future work, we will use these new models to extract convolutional feature maps and seek further improvement.}
\label{tab:state-of-the-art}
\end{table}

\section{Conclusions} \label{conclusion}

In this paper, we propose Evolution-Preserving Trajectory (EPT) descriptors, which integrate the benefits of dense trajectories, deep-learning features, and rank pooling. Deep architectures are utilized to extract discriminative convolutional feature maps, and rank pooling is applied to each trajectory, preserving the temporal evolution of appearance and motion along the trajectory. The EPT descriptor significantly outperforms the previous state-of-the-art trajectory-based descriptors. Combining EPT descriptors that preserve short-term evolution and the video-wide rank pooling that captures long-term dynamics, we are able to advance the  state-of-the-art performance on Hollywood2 dataset.

{\small
\bibliographystyle{abbrvnat}
\bibliography{shortstrings,pubs2,pubs,egbib}

\begin{thebibliography}{29}
\providecommand{\natexlab}[1]{#1}
\providecommand{\url}[1]{\texttt{#1}}
\expandafter\ifx\csname urlstyle\endcsname\relax
  \providecommand{\doi}[1]{doi: #1}\else
  \providecommand{\doi}{doi: \begingroup \urlstyle{rm}\Url}\fi

\bibitem[Bilen et~al.(2016)Bilen, Fernando, Gavves, Vedaldi, and
  Gould]{bilen2016dynamic}
H.~Bilen, B.~Fernando, E.~Gavves, A.~Vedaldi, and S.~Gould.
\newblock Dynamic image networks for action recognition.
\newblock In \emph{Proc. CVPR}, 2016.

\bibitem[Dalal and Triggs(2005)]{Dalal-Triggs-CVPR05}
N.~Dalal and B.~Triggs.
\newblock Histograms of oriented gradients for human detection.
\newblock In \emph{Proc. CVPR}, 2005.

\bibitem[Doll\'ar et~al.(2005)Doll\'ar, Rabaud, Cottrell, and
  Belongie]{Dollar-et-al-VSPETS05}
P.~Doll\'ar, V.~Rabaud, G.~Cottrell, and S.~Belongie.
\newblock Behavior recognition via sparse spatio-temporal features.
\newblock In \emph{ICCV Workshop on Visual Surveillance \& Performance
  Evaluation of Tracking and Surveillance}, 2005.

\bibitem[Fernando et~al.(2015)Fernando, Gavves, M., Ghodrati, and
  Tuytelaars]{Fernando-etal-CVPR15}
B.~Fernando, E.~Gavves, J.~O. M., A.~Ghodrati, and T.~Tuytelaars.
\newblock Modeling video evolution for action recognition.
\newblock In \emph{Proc. CVPR}, 2015.

\bibitem[Fernando et~al.(2016)Fernando, Anderson, Hutter, and
  Gould]{fernando2016discriminative}
B.~Fernando, P.~Anderson, M.~Hutter, and S.~Gould.
\newblock Discriminative hierarchical rank pooling for activity recognition.
\newblock In \emph{Proc. CVPR}, 2016.

\bibitem[Girshick et~al.(2014)Girshick, Donahue, Darrell, and
  Malik]{Girshick-et-al-CVPR14}
R.~Girshick, J.~Donahue, T.~Darrell, and J.~Malik.
\newblock Rich feature hierarchies for accurate object detection and semantic
  segmentation.
\newblock In \emph{Proc. CVPR}, 2014.

\bibitem[He et~al.(2016)He, Zhang, Ren, and Sun]{he2015deep}
K.~He, X.~Zhang, S.~Ren, and J.~Sun.
\newblock Deep residual learning for image recognition.
\newblock \emph{Proc. CVPR}, 2016.

\bibitem[Hoai and Zisserman(2014)]{Hoai-Zisserman-ACCV14}
M.~Hoai and A.~Zisserman.
\newblock Thread-safe: Towards recognizing human actions across shot
  boundaries.
\newblock In \emph{Proc. ACCV}, 2014.

\bibitem[Jain et~al.(2015)Jain, van Gemert, and Snoek]{jain201515}
M.~Jain, J.~C. van Gemert, and C.~G. Snoek.
\newblock What do 15,000 object categories tell us about classifying and
  localizing actions?
\newblock In \emph{Proc. CVPR}, 2015.

\bibitem[Jiang et~al.(2014)Jiang, Liu, Zamir, Toderici, Laptev, Shah, and
  Sukthankar]{jiang2014thumos}
Y.~Jiang, J.~Liu, A.~R. Zamir, G.~Toderici, I.~Laptev, M.~Shah, and
  R.~Sukthankar.
\newblock Thumos challenge: Action recognition with a large number of classes.
\newblock In \emph{ECCV Workshop}, 2014.

\bibitem[Krizhevsky et~al.(2012)Krizhevsky, Sutskever, and
  Hinton]{Krizhevsky-et-al-NIPS12}
A.~Krizhevsky, I.~Sutskever, and G.~Hinton.
\newblock {ImageNet} classification with deep convolutional neural networks.
\newblock In \emph{NIPS}, 2012.

\bibitem[Laptev(2005)]{Laptev-IJCV05}
I.~Laptev.
\newblock On space-time interest points.
\newblock \emph{IJCV}, 64\penalty0 (2--3):\penalty0 107--123, 2005.

\bibitem[LeCun et~al.(1989)LeCun, Boser, Denker, and
  Henderson]{LeCun-et-al-NC89}
Y.~LeCun, B.~Boser, J.~S. Denker, and D.~Henderson.
\newblock Backpropagation applied to handwritten zip code recognition.
\newblock \emph{Neural Computation}, 1\penalty0 (4):\penalty0 541--551, 1989.

\bibitem[Lev et~al.(2016)Lev, Sadeh, Klein, and Wolf]{lev2015rnn}
G.~Lev, G.~Sadeh, B.~Klein, and L.~Wolf.
\newblock Rnn fisher vectors for action recognition and image annotation.
\newblock In \emph{Proc. ECCV}, 2016.

\bibitem[Marszalek et~al.(2009)Marszalek, Laptev, and
  Schmid]{Marszalek-et-al-CVPR09}
M.~Marszalek, I.~Laptev, and C.~Schmid.
\newblock Actions in context.
\newblock In \emph{Proc. CVPR}, 2009.

\bibitem[Perronnin et~al.(2010)Perronnin, S\'anchez, and
  Mensink]{Perronnin-et-al-ECCV10}
F.~Perronnin, J.~S\'anchez, and T.~Mensink.
\newblock Improving the fisher kernel for large-scale image classification.
\newblock In \emph{Proc. ECCV}, 2010.

\bibitem[Russakovsky et~al.(2015)Russakovsky, Deng, Su, Krause, Satheesh, Ma,
  Huang, Karpathy, Khosla, Bernstein, Berg, and
  Fei-Fei]{Russakovsky-etal-IJCV15}
O.~Russakovsky, J.~Deng, H.~Su, J.~Krause, S.~Satheesh, S.~Ma, Z.~Huang,
  A.~Karpathy, A.~Khosla, M.~Bernstein, A.~C. Berg, and L.~Fei-Fei.
\newblock Imagenet large scale visual recognition challenge.
\newblock \emph{IJCV}, 115\penalty0 (3):\penalty0 211--252, 2015.

\bibitem[Simonyan and Zisserman(2014)]{Simonyan-Zisserman-NIPS14}
K.~Simonyan and A.~Zisserman.
\newblock Two-stream convolutional networks for action recognition in videos.
\newblock In \emph{NIPS}, 2014.

\bibitem[Simonyan and Zisserman(2015)]{simonyan2014very}
K.~Simonyan and A.~Zisserman.
\newblock Very deep convolutional networks for large-scale image recognition.
\newblock In \emph{Proc. ICLR}, 2015.

\bibitem[Soomro et~al.(2012)Soomro, Zamir, and Shah]{Soomro-et-al-TR12}
K.~Soomro, A.~R. Zamir, and M.~Shah.
\newblock {UCF101}: A dataset of 101 human action classes from videos in the
  wild.
\newblock Technical Report CRCV-TR-12-01, University of Central Florida, 2012.

\bibitem[Tran et~al.(2015)Tran, Bourdev, Fergus, Torresani, and
  Paluri]{tran2015learning}
D.~Tran, L.~Bourdev, R.~Fergus, L.~Torresani, and M.~Paluri.
\newblock Learning spatiotemporal features with 3d convolutional networks.
\newblock In \emph{Proc. ICCV}. IEEE, 2015.

\bibitem[Vapnik(1998)]{Vapnik-98}
V.~Vapnik.
\newblock \emph{Statistical Learning Theory}.
\newblock Wiley, New York, NY, 1998.

\bibitem[Wang and Schmid(2013)]{Wang-Schmid-ICCV13}
H.~Wang and C.~Schmid.
\newblock Action recognition with improved trajectories.
\newblock In \emph{Proc. ICCV}, 2013.

\bibitem[Wang et~al.(2011)Wang, Klaser, Schmid, and Liu]{Wang-et-al-CVPR11}
H.~Wang, A.~Klaser, C.~Schmid, and C.-L. Liu.
\newblock Action recognition by dense trajectories.
\newblock In \emph{Proc. CVPR}, 2011.

\bibitem[Wang et~al.(2015{\natexlab{a}})Wang, Qiao, and Tang]{wang2015action}
L.~Wang, Y.~Qiao, and X.~Tang.
\newblock Action recognition with trajectory-pooled deep-convolutional
  descriptors.
\newblock In \emph{Proc. CVPR}, 2015{\natexlab{a}}.

\bibitem[Wang et~al.(2016)Wang, Xiong, Wang, Qiao, Lin, Tang, and
  Van~Gool]{wang2016temporal}
L.~Wang, Y.~Xiong, Z.~Wang, Y.~Qiao, D.~Lin, X.~Tang, and L.~Van~Gool.
\newblock Temporal segment networks: Towards good practices for deep action
  recognition.
\newblock In \emph{Proc. ECCV}, pages 20--36. Springer, 2016.

\bibitem[Wang et~al.(2015{\natexlab{b}})Wang, Cao, Shen, Liu, and
  Shen]{wang2015temporal}
P.~Wang, Y.~Cao, C.~Shen, L.~Liu, and H.~T. Shen.
\newblock Temporal pyramid pooling based convolutional neural networks for
  action recognition.
\newblock \emph{arXiv preprint arXiv:1503.01224}, 2015{\natexlab{b}}.

\bibitem[Wang and Hoai(2016)]{Wang-Hoai-CVPR16}
Y.~Wang and M.~Hoai.
\newblock Improving human action recognition by non-action classification.
\newblock In \emph{Proc. CVPR}, 2016.

\bibitem[Willems et~al.(2008)Willems, Tuytelaars, and
  Gool]{Willems-et-al-ECCV08}
G.~Willems, T.~Tuytelaars, and L.~V. Gool.
\newblock An efficient dense and scale-invariant spatio-temporal interest point
  detector.
\newblock In \emph{Proc. ECCV}, 2008.

\end{thebibliography}
}

\end{document}